# LABEL-DRIVEN WEAKLY-SUPERVISED LEARNING FOR MULTIMODAL DEFORMABLE IMAGE REGISTRATION

*Yipeng Hu[1,2], Marc Modat[1], Eli Gibson[1], Nooshin Ghavami[1], Ester Bonmati[1], Caroline M. Moore[3], Mark Emberton[3], J. Alison Noble[2], Dean C. Barratt[1,4], Tom Vercauteren[1,4]*

[1] Centre for Medical Image Computing, University College London, UK
[2] Institute of Biomedical Engineering, University of Oxford, UK
[3] Division of Surgery and Interventional Science, University College London, London, UK
[4] Wellcome / EPSRC Centre for Interventional and Surgical Sciences, University College London, UK

**ABSTRACT**

Spatially aligning medical images from different modalities remains a challenging task, especially for intraoperative applications that require fast and robust algorithms. We propose a weakly-supervised, label-driven formulation for learning 3D voxel correspondence from higher-level label correspondence, thereby bypassing classical intensity-based image similarity measures. During training, a convolutional neural network is optimised by outputting a dense displacement field (DDF) that warps a set of available anatomical labels from the moving image to match their corresponding counterparts in the fixed image. These label pairs, including solid organs, ducts, vessels, point landmarks and other *ad hoc* structures, are only required at training time and can be spatially aligned by minimising a cross-entropy function of the warped moving label and the fixed label. During inference, the trained network takes a new image pair to predict an optimal DDF, resulting in a fully-automatic, label-free, real-time and deformable registration. For interventional applications where large global transformation prevails, we also propose a neural network architecture to jointly optimise the global- and local displacements. Experiment results are presented based on cross-validating registrations of 111 pairs of T2-weighted magnetic resonance images and 3D transrectal ultrasound images from prostate cancer patients with a total of over 4000 anatomical labels, yielding a median target registration error of 4.2 mm on landmark centroids and a median Dice of 0.88 on prostate glands.

## 1. INTRODUCTION

Intraoperative transrectal ultrasound (TRUS) images are used to guide the majority of targeted biopsies and focal therapies for prostate cancer patients but generally displays poor contrast between the healthy- and cancerous regions. Multi-parametric magnetic resonance (MR) images have been shown clinically important in detecting and localising prostate cancers, but are usually acquired outside of interventions [1]. In general, multimodal image registration aims to fuse the preoperative planning information such as the tumour locations, obtained from diagnostically purposed imaging, with real-time but often low quality intraoperative imaging [2]. It thus may substantially improve patient care [3] and refine disease risk stratification [4].

However, like many other ultrasound-guided medical procedures, this application represents a typical example where no robust image similarity measure has been demonstrated. For instance, anatomically different structures, such as prostate inner-outer gland separation defined on TRUS [5] and central-peripheral zonal boundary visible on MR, possess strong statistical correlation between them. This leads to false alignment using most, if not all, of the established intensity-based similarity measures and the associated registration methodologies. On the other hand, feature-based image registration methods proposed for this application rely on relatively sparse features that must be identified consistently between two imaging modalities, such as prostate capsule surface. Obtaining such manual landmarks and/or segmentations from both images for each patient is time-consuming, user-dependent and often infeasible intraoperatively. Alternative methods using automated hand-engineered image features can be highly sensitive to data, initialisation and often needs strong application-specific prior regularisation, e.g. [6, 7].

Recent machine-learning-based methods, particularly using neural networks, have been proposed to replace and accelerate the registration algorithms based on iterative optimisation with acquired knowledge from population data, e.g. [8, 9], or to provide better constrained transformation models [10]. However, most still rely on the choice of image similarity measure or on non-trivial, and potentially non-realistic, application-specific assumptions in procuring large number of ground-truth transformations for training, such as those from simulations [11, 12], existing registration methods [13] or manual rigid alignment [14].

In this work, considering applications with no known robust image similarity measure, we propose a novel neural-network-based learning approach, inferring dense voxel correspondence from all types of identifiable labels representing correspondent anatomical structures. In the example application, we describe a specific network architecture to effectively predict a composite of global- and local deformation encountered during the prostate cancer interventions, registering MR- and TRUS images.

## 2. METHOD

### 2.1. Label-Driven Correspondence Learning

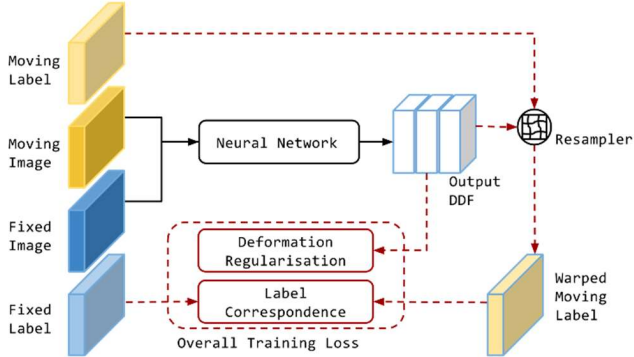

**Fig.1.** Illustration of the training process of the proposed label-driven registration framework (described in Section 2.1), where the red-dotted lines represent data flows only required in training.

Image registration requires establishing spatially corresponding voxels between a pair of *moving-* and *fixed images*. From a machine learning perspective, ground-truth voxel-level labels for correspondence learning are scarce and, in most scenarios, impossible to reliably obtain from medical image data. However, higher-level corresponding structures are much more feasible to annotate with anatomical knowledge, such as the same organ, lesion or other structures appearing in the two images. In this section, we introduce a generalisable framework to use these images and labels as training data, to enable an automatic, deformable image registration that only requires image data during inference.

Given a pair of training images to align, a neural network computes a dense displacement field (DDF) to indicate the correspondence: a DDF-warped *moving label* voxel corresponds to the *fixed label* voxel at the same image coordinate, if they both are foreground (ones) or background (zeros). Each label pair in training is considered to serve as a weak label of voxel correspondence at the spatial location, which is measured by a two-class cross-entropy function.

In medical image applications, different cases may have different types and numbers of corresponding structures. Therefore, we propose to feed different label pairs randomly into training, one corresponding pair at each iteration for each image pair, so that a minibatch optimisation scheme takes an equal number of image- and label pairs in each minibatch. Considering all the label pairs given an image pair being the conditionally independent data of the likelihood function representing voxel correspondence, the gradient computed from this minibatch is still an unbiased estimator of the gradient of the loss function using the entire training set, as in stochastic gradient descent [15].

Furthermore, large number of manual labels are in general prone to errors and can be difficult to verify thoroughly. To mitigate labelling uncertainty, we sample higher confidence labels more frequently (as described in Section 3). This weights the objective function towards the subset of labels having the highest quality through expensive expert labelling or multiple observer consensus.

As shown in Figure 1, in training, the label correspondence in terms of cross-entropy is computed between the warped moving label and the fixed label, neither of which are used as input to the network. Therefore they are not required in inference, i.e. registration. The moving- and fixed images are used only as inputs to the neural network without directly contributing to the loss function. Therefore, no explicit image similarity measure needs to be enforced. Depending on the availability of label data, smoothness of the DDFs is added to the overall loss function, such as bending energy or $L^2$-norm of the displacement gradients, in addition to the network architectural regularisation. The labels (and the images if required in inference) are warped with an image resampler, such as linear-, cubic- or spline interpolation.

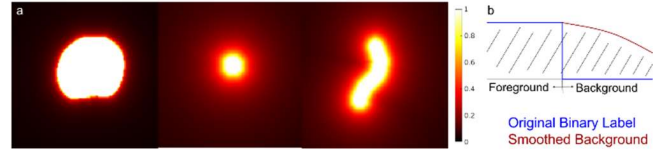

**Fig.2.** a: Example slices of the spatially smoothed label maps for different types of landmarks, a prostate gland, a cyst and a urethra, from left to right; b: illustration of the profile of a smoothed label maps with the shaded area indicating the normalised sum $M$ of the map. Details are described in Section 2.1.

Direct use of the binary masks may cause an over-fitted and under-regularised network [16], independent of the class balance issue. As illustrated in Figure 2b, we propose a one-sided label smoothing mechanism to 1) provide a spatially smoothed probability map on the original background region while the original foreground region remains associated with one-hot probability vectors; and 2) without affecting the one-hot probabilities in the foreground region, normalise the sum of the map (i.e. the weight of the shaded area in Figure 2b) so that all labels in a given image are associated with the same weight. Such a smoothing is achieved by an element-wise nonlinear function with a parameter controlling the area under the curve to reach a subscribed sum $M$ (here, $M$ equals to the sum over all voxels of an inverse distance transform, measured in voxel units, of the label with largest volume, i.e. full gland segmentation). While the mapping function should be ideally designed based on the underlying correspondence distribution [16], a heuristically motivated function is used in this work: $p_i = 1 - (1 - d_i^{-1})^x$, where $d_i^{-1}$ is computed from inverse distance transform, measured in voxel units, at voxel $i$ in the background region, and $x$ is a re-weighting parameter, so that $\sum_i p_i = M$; $p_i=1$ for the foreground region. Figure 2a provides examples of these label maps $p_i$ used as input probabilities to the cross-entropy loss. From the optimisation perspective, this provides a larger capture range and more balanced weighting between labels from the same image, as opposed to a simple Gaussian smoothing.

**2.2. Learning Interventional Deformation**

In this section, we describe an instantiated implementation of the described framework, for registering the prostate MR-to TRUS images described in Section 1. In intraoperative applications of this kind, considerable spatial difference in physical space is common between preoperative- and intraoperative images, caused by different imaging reference coordinate systems, imaging sampling parameters and patient movement. It is usually represented by a global rigid- or affine transformation [17], which could remain non-penalised under typical regularisation strategies, such as the bending energy. Therefore, a network directly predicting regularised DDFs should in theory be able to model both local- and global deformations. In practice, however, the regions under-supported by available labels may receive diminishing and/or perturbing gradients that are sensitive to the weighting between the cross-entropy loss and the regularisation. For example, most landmarks are defined within or near to the prostate gland in our application: in the initial experiments, directly predicting DDFs yielded implausible deformations and large target registration errors (TREs), when testing landmarks were outside the gland (see the results in Section 4). To combat this practical issue, we augment the network with a soft architectural constraint to encourage the learning of the global transformation, as illustrated in Figure 3.

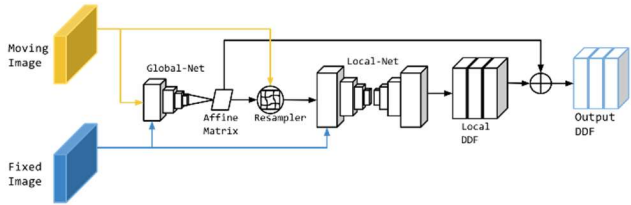

**Fig.3.** Illustration of the inference part of the proposed registration network (described in Section 2.2) for interventional applications.

First, the moving image is linearly resized to the same size as the fixed image. After receiving the concatenated image pair, a sub-network *global-net* predicts a 12 degrees of freedom affine transformation matrix. The original moving image is warped onto the fixed image coordinates by the global affine transformation and the result is concatenated with the original fixed image. A second sub-network *local-net* then takes the concatenation as input to generate a non-rigid local DDF regularised by a weighted bending energy [17]. The affine transformation and the local DDF are then composed to produce the final output DDF to warp the moving label to match the fixed label using a trilinear image resampler. It is not guaranteed that the learned local DDFs do not contain global transformation, but our approach is designed to improve the learning by preliminarily warping the moving image, also suggested in the previous work [18], and converges faster in practice.

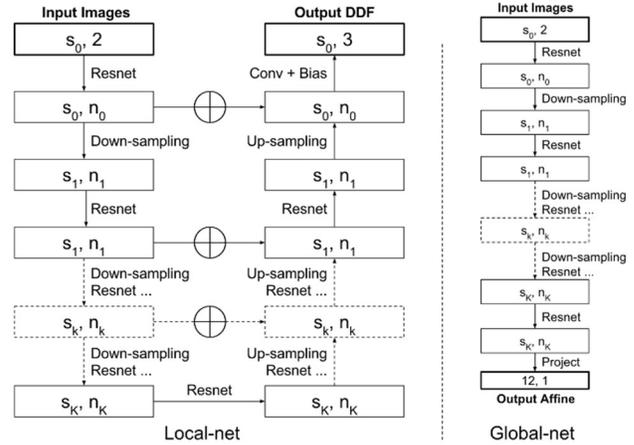

**Fig.4.** Illustration of the network architecture of the local-net (left), $s_k$ and $n_k$ (k=1, …, K), are the sizes and numbers of channels of feature maps, with K=4 and $s_0$ being the fixed image size. The global-net (right) shares the same down-sampling architecture with a linear projection to an affine matrix. Initial $n_0$ are 32 and 4 for local-net and global-net, respectively. See details in Section 2.2.

Following our previous work in segmenting prostate gland from TRUS images [19] and the proposed method for learning optical flow [20], the local-net is designed as a 3D convolutional neural network with four down-sampling blocks, followed by four up-sampling blocks, with four summation operations shortcutting the network at different resolution levels. As illustrated in Figure 4, it begins with $n_0=32$ initial channels of feature maps, successively doubles the number of channels and halves the feature map size with the down-sampling blocks, and *vice versa* with the up-sampling blocks. Each of these blocks consists of a residual network (Resnet) unit shortcutting two convolutional- and batch normalisation layers with rectified linear units. Down-sampling and up-sampling are achieved by convolution and transpose convolution, respectively, both with strides of two. Predicting spatial transformation is sensitive to network initialisation [21]. For initialising the output so that the resampled labels generate meaningful initial gradients, the output layer is an additional convolution (Conv) added to a bias term, without batch normalisation or nonlinear activation. It enables random initialisation with zero mean and a small variation on the final layer.

The global-net shares the same architecture (using independently learnable parameters) as the four down-sampling blocks of the local-net, with a smaller number of $n_0=4$ initial channels. The last convolutional layer is linearly projected by a learnable fully-connected matrix to a vector containing 12 affine transformation parameters. Similarly, this allows a random initialisation centred at an identity matrix with zero translation.

## 3. EXPERIMENT

A total of 111 pairs of T2-weighted MR- and TRUS images from 76 patients were acquired during SmartTarget® clinical trials [22]. Each patient has up to three image data sets due to multiple procedures he entered, i.e. biopsy and therapy, or multiple ultrasound volumes being acquired at the beginning and the conclusion of a procedure according to the therapy trial protocol. A range of 35 - 112 parasagittal TRUS frames were acquired in each case by rotating a digital stepper with recorded relative angles covering the majority of the prostate gland, and then used to reconstruct a 3D volume in Cartesian coordinates. Both MR- and TRUS images were normalised to zero-mean with unit-variance intensities after being resampled to 1.0 mm$^3$ isotropic voxels.

From these patients, a total of 2351 pairs of corresponding anatomical landmarks were labelled by two medical imaging research fellows and a research student using an in-house voxel-painting tool on the original image data. Besides full gland segmentations for all cases, these landmarks include apex, base, urethra, visible lesions, junctions between the gland, vas deference and the seminal vesicles, and many other patient-specific point landmarks such as calcifications and fluid-filled cysts. Among these, 330 pairs double-checked and confirmed by a second observer were considered as ones with high confidence. Prostate gland segmentations were all considered of high confidence. The labels were resampled to the same sizes of their respective images before being smoothed by the normalised inverse distance transform (Section 2.1).

The method was implemented in TensorFlow™ with the 3D resampler module and a 3D image augmentation layer adapted from the open-source code in NiftyNet [23]. Each network was trained on a 24GB NVIDIA® Quadro™ P6000 for 12 hours, using an Adam optimiser starting at a learning rate of $10^{-5}$ with a minibatch size of 10. The bending energy weight and weight-decay were set to $10^{-2}$ and $10^{-6}$, respectively. The global-net was trained first for 1000 iterations. The high confidence labels were sampled twice as more frequently as the others in training.

For comparison, the global-net and the local-net were trained separately by minimising the same loss function. In the case of training the local-net alone, the originally affine-resampled moving image (in Figure 3) was linearly resized to concatenate with the fixed image. All the applicable hyper-parameters were set to the same values.

In each fold of the 10-fold patient-level cross-validation, test data from 7-8 patients were held out while the data from the remainder patients were used in training. Two measures are reported in this study, TREs on centroids of the landmarks with high confidence and Dice scores on the prostate glands. These two independently-calculated metrics on left-out test data directly relate to two of the clinical requirements in the registration-enabled guidance, avoiding surrounding structures and locating regions of interest.

## 4. RESULTS

More than 4 3D registrations per second can be performed on the same GPU. Figure 5 shows example slices from the input MR-TRUS image pairs and the registered MR images. A median TRE (5$^{th}$ - 95$^{th}$ percentiles) of 4.2 (1.0 - 14.9) mm on landmark centroids and a median Dice of 0.88 (0.80 - 0.94) on prostate glands were achieved. The median TRE and Dice were 9.4 (3.0 - 20.4) mm and 0.73 (0.46 - 0.87) from the global-net and, 8.5 (1.5 - 22.7) mm and 0.86 (0.46 - 0.91) from the local-net, respectively. The TREs from the proposed network are significantly better than those from the alternative global-net or local-net alone (with both *p-values* < 0.001 from paired t-tests), while the competitive Dice from the local-net may suggest an over-fitting to the data around prostate gland areas with the unaided local-net, also more susceptible to visually non-realistic deformations.

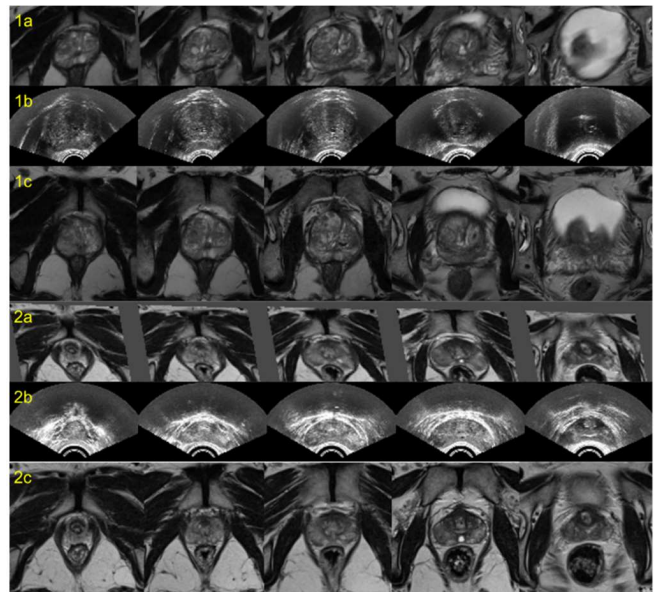

**Fig.5.** Example image slices from two cases, 1 and 2. Rows a, b and c contain slices from the registered moving MR images, the fixed TRUS images and visually closest slices from original MR images for comparison, respectively

## 5. CONCLUSION

We have introduced a flexible framework that can utilise different image resamplers, deformation regularisers and all types of anatomical labels, enabling a fully-automatic multimodal image registration algorithm using only input image pair. We have also described a network architecture, effectively learning global- and local interventional deformations. Registration results are reported from a rigorous validation on 111 pairs of labelled intraoperative prostate image data. Future research aims to investigate the generalisation of the proposed method to data from different centres and to a wider range of applications.